\documentclass[11pt]{article}
\usepackage{amsmath,amssymb,amsfonts}
\usepackage{graphicx}
\usepackage{hyperref}
\usepackage{natbib}
\usepackage{url}
\usepackage{fullpage}

\usepackage{pgfplots}
\usepackage{caption}

\pgfplotsset{compat=1.18}

\usepackage{tikz}
\usetikzlibrary{positioning}
\usetikzlibrary{calc}
\usetikzlibrary{arrows.meta}
\usetikzlibrary{decorations.pathreplacing}

\title{Beyond Words: A Latent Memory Approach \\ to Internal Reasoning in LLMs}
\author{José I. Orlicki \\
Independent Researcher\\
\texttt{josepreprints@gmail.com}}
\date{}

\begin{document}

\maketitle

\begin{abstract}
Recent advances in large language models (LLMs) have popularized the \emph{chain-of-thought} (CoT) paradigm, in which models produce explicit reasoning steps in natural language. Although this approach improves interpretability and facilitates external auditing, it may not represent the most computationally efficient method for internal reasoning. In contrast, human cognition relies on implicit mental representations that recall past sensory and episodic information without requiring complete verbalization. In this paper, we propose a framework that integrates \emph{implicit mental representations} into the internal reasoning processes of LLMs. Preliminary experiments indicate that incorporating an Implicit Memory Module (IMM) into a simple GPT model yields a reduction of between 35\% and 57\% in final training loss compared to a regular GPT baseline. The addition of an explicit interpretability channel (e.g., a chain-of-thought decoder) is straightforward to implement within this approach. We outline theoretical foundations, propose technical mechanisms to scale the memory module, and discuss how these ideas may lead to more efficient and robust reasoning, with optional future extensions for explicit auditability.
\end{abstract}

\section{Introduction}
The chain-of-thought (CoT) approach in large language models has demonstrated that explicitly generated intermediate reasoning steps can improve performance on complex tasks \citep{wei2022chain}. However, explicit reasoning not only incurs computational overhead, but may also diverge from what would be the most efficient internal process. Human cognition, for example, relies on implicit mental representations to recall episodic and sensory information without fully articulating every internal state \citep{baddeley2003working}. 

In this paper, we propose incorporating \emph{implicit mental representations}—latent internal states that capture past context and sensory-like memory—into the reasoning processes of LLMs. Our approach centers on an Implicit Memory Module (IMM) that dynamically stores and retrieves latent representations. While we acknowledge the benefits of an explicit interpretability channel (such as a chain-of-thought decoder), here we focus on the latent mechanisms; the explicit channel is noted as a promising avenue for future work.

\section{Background}

\subsection{Explicit versus Implicit Reasoning} Current CoT methods force models to articulate their reasoning in human-readable language, aiding in auditing but introducing extra computational costs and inefficiencies. Explicit reasoning requires generating verbose explanations, which can distract from the core internal computations and limit scalability—especially when handling complex tasks where the model might benefit from more compact internal representations. In contrast, human memory leverages implicit representations: when recalling a past event, we reassemble sensory impressions and contextual cues without a complete verbal account \citep{baddeley2003working}. For example, we remember a familiar melody without having to recite every single note. This limitation of explicit representations motivates the development of internal reasoning mechanisms that rely primarily on latent states, reserving explicit verbalization only for situations where interpretability is essential.

\subsection{Latent Representations in LLMs}
Several influential works have explored the nature of latent representations in language models. \citet{petroni2019language} demonstrated that LLMs can act as implicit knowledge bases, storing factual information in their latent spaces. Similarly, \citet{wei2022architecture} investigated how different model architectures and pretraining objectives impact zero-shot generalization, showing that improved latent representations correlate with enhanced generalization capabilities. A mathematical framework for transformer circuits \citep{anthropic2022transformer} further elucidates how information is organized and processed across layers, with lower layers capturing syntactic patterns and higher layers encoding complex semantics.

Key insights include:
\begin{itemize}
    \item \textbf{Specialized Directions:} LLMs develop latent directions that encode specific concepts or linguistic features.
    \item \textbf{Layerwise Abstraction:} Lower layers capture syntactic details while higher layers integrate complex semantic and world knowledge.
    \item \textbf{Scaling Benefits:} The quality of latent representations improves with increased model scale and optimized architectures.
\end{itemize}

\section{Proposed Framework}
\subsection{Overview}
We propose an architectural augmentation where an LLM is equipped with an \emph{implicit memory module} (IMM). This module stores latent representations generated during processing and retrieves relevant information when needed for internal reasoning. Our focus is on efficiently using these latent mechanisms to enhance performance. One could also readily add an explicit CoT decoder to provide a human-interpretable narrative of the internal reasoning process.

\subsection{Mathematical Formulation}

Let $\{h_1, h_2, \dots, h_T\}$ denote the sequence of hidden states produced by a Transformer model for a given input. We introduce a memory bank $M \in \mathbb{R}^{N \times d}$, where $N$ is the number of memory slots and $d$ is the hidden state dimensionality.

\paragraph{Memory Write:} At selected time steps $t$, the model writes a summary representation $s_t$ into $M$:
\[
    s_t = f_{\text{write}}(h_t), \quad M[i] \leftarrow s_t \quad \text{for some } i \in \{1,\dots,N\},
\]
where $f_{\text{write}}$ is a learned projection. In other words, $f_{\text{write}}$ is parameterized by trainable weights and is optimized jointly with the rest of the network during training, ensuring that it effectively compresses the relevant information from $h_t$.

\paragraph{Memory Read:} The model computes a query vector $q_t$ from the current hidden state:
\[
    q_t = f_{\text{query}}(h_t),
\]
with $f_{\text{query}}$ also implemented as a learnable transformation. This means that $f_{\text{query}}$ is similarly trained to extract the most pertinent features of $h_t$ to retrieve information from the memory bank via attention:
\[
    \alpha = \text{softmax}(M\, q_t^\top), \quad r_t = \sum_{i=1}^{N} \alpha_i M[i].
\]
By jointly optimizing these components, the model adapts its internal representations to best support memory integration and retrieval throughout the reasoning process.

\subsection{Integrating Implicit Memory with Optional Explicit Oversight}
We envision a dual-process model comprising:
\begin{enumerate}
    \item \textbf{Implicit Reasoning:} The primary process wherein the IMM dynamically stores and retrieves latent information to support reasoning.
    \item \textbf{Optional Explicit Oversight:} A lightweight, auxiliary decoder may be incorporated to generate explicit chains-of-thought if interpretability is required. In our current work, however, we emphasize the latent pathway and reserve full explicit auditing as potential future work.
\end{enumerate}
This dual-process design offers efficiency in computation while keeping the door open for enhanced transparency if needed.

\section{Implementation Details}
This section describes the architectural design, training dynamics, and procedural aspects of our model augmented with an IMM.

\subsection{Model Architecture}
\begin{figure}[ht]
    \centering
    \resizebox{\textwidth}{!}{ % Resizes diagram to fit within page width
    \begin{tikzpicture}[
        node distance=1.5cm and 2cm,
        auto,
        block/.style={draw, rectangle, rounded corners, minimum height=1cm, minimum width=3cm, align=center},
        arrow/.style={-{Latex[round]}, thick}
    ]

    % Input and Embedding
    \node[block, fill=blue!10] (input) {Input Tokens};
    \node[block, fill=blue!10, right=of input] (embed) {Embedding Layer};

    % Transformer with IMM
    \node[block, fill=green!10, right=of embed, minimum width=4cm, minimum height=2.2cm] (transformer) {Transformer Layers \\ with Implicit Memory Module (IMM)};
      
    % Memory Bank (below transformer)
    \node[block, fill=gray!20, below=0.6cm of transformer] (mem) {Memory Bank};

    % Output branch
    \node[block, fill=red!10, right=of transformer] (output) {Output Tokens};

    % Optional Explicit Oversight (for future extension)
    \node[block, fill=yellow!10, below=of mem] (explicit) {CoT Decoder};

    % Arrows connecting nodes
    \draw[arrow] (input) -- (embed);
    \draw[arrow] (embed) -- (transformer);
    \draw[arrow] (transformer) -- (output);

    % Memory interactions
    \draw[arrow, <->] (transformer.south) -- (mem.north);
    \draw[arrow] (mem.south) -- (explicit.north);
    \draw[arrow, bend left=20] (explicit.east) to (output.south);

    % Labels
    \node[above=0.2cm of embed] {1. Embedding};
    \node[above=0.2cm of transformer] {2. Transformer + IMM};
    \node[above=0.2cm of output] {3. Output Generation};
    \node[left=0.2cm of explicit] {4. Optional Explicit Decoder};

    \end{tikzpicture}
    }
    \caption{Model architecture showing the Transformer layers with the Implicit Memory Module (IMM) and an optional explicit decoder for future interpretability enhancements.}
    \label{fig:architecture}
\end{figure}
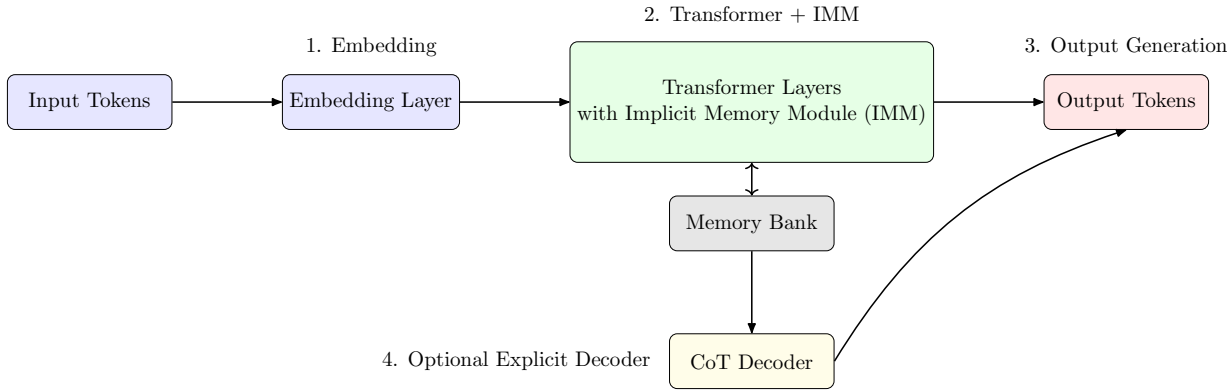

Our approach augments a standard Transformer with an IMM that functions as a differentiable key-value memory. The IMM stores and retrieves latent representations that are integrated into the hidden states, thereby enhancing internal reasoning. Although one could attach an explicit decoder for chain-of-thought outputs, in this work we focus on the IMM. Figure~\ref{fig:architecture} illustrates the architecture, with the optional explicit channel noted as a possible extension rather than a core component.

Initially, we experimented with a memory module that utilized a fully linear structure with square dimensions on embeddings size ($n\_embd$), allowing the memory mechanism to be fully parameterized in high-dimensional space. The initial implementation employed 16 memory slots, each fully connected to the embedding dimensions, providing a dense representation of latent knowledge. However, for scalability purposes, we introduced an optimized variant by constraining one dimension to a fixed projection, akin to Linformer-style compression. 

This resulted in the transition from our initial implementation to a \texttt{Linformer} style module, which significantly reduces computational complexity from $O(n\_embd^2)$ to $O(n\_embd \cdot k)$, where $k$ is the low-rank projection size. Furthermore, to maintain adaptability across different model sizes, we dynamically scale the number of memory slots based on the embedding dimension using the heuristic $\texttt{num\_slots} = \sqrt{n\_embd}$. This strategy ensures a balanced trade-off between expressiveness and computational efficiency, allowing the memory module to scale effectively with larger transformer architectures while avoiding quadratic complexity bottlenecks.

\subsection{IMM Training and Inference Dynamics}

\begin{figure}[ht]
    \centering
    \resizebox{\textwidth}{!}{
    \begin{tikzpicture}[
        node distance=2cm and 3cm,
        auto,
        block/.style={draw, rectangle, rounded corners, minimum height=1cm, minimum width=3cm, align=center},
        arrow/.style={-{Latex[round]}, thick},
        dashedarrow/.style={-{Latex[round]}, thick, dashed}
    ]
    
    % Input hidden state
    \node[block, fill=green!10] (inputHT) {$h_t$\\(Input Hidden State)};
    
    % Write operation: computing summary representation
    \node[block, fill=orange!20, right=of inputHT] (writeOp) {$s_t = f_{\text{write}}(h_t)$\\(Memory Write)};
    
    % Memory Bank
    \node[block, fill=gray!20, below=of writeOp] (memBank) {Memory Bank $M$};
    
    % Query computation from h_t
    \node[block, fill=orange!20, below=of inputHT] (query) {$q_t = f_{\text{query}}(h_t)$\\(Query Vector)};
    
    % Read operation: retrieving memory using the query
    \node[block, fill=orange!20, right=of query] (readOp) {$r_t = \sum \alpha_i M[i]$\\(Memory Read via Attention)};
    
    % Integration: combining h_t with transformed r_t and normalizing
    \node[block, fill=green!10, right=of readOp] (integrate) {$\tilde{h}_t = \text{LayerNorm}\big(h_t + g(r_t)\big)$\\(Updated Hidden State)};
    
    % Arrows for Write Path
    \draw[arrow] (inputHT.east) -- node[above] {Write} (writeOp.west);
    \draw[arrow] (writeOp.south) -- (memBank.north);
    
    % Arrows for Read Path
    \draw[arrow] (inputHT.south) -- node[left] {Compute Query} (query.north);
    \draw[arrow] (query.east) -- (readOp.west);
    
    % Memory Read via Attention arrow
    \draw[arrow] (memBank.east) to[out=-90, in=-90] node[right, xshift=2mm, yshift=-4mm] {Attention} (readOp.south);
    
    % Integration via Skip Connection
    \draw[arrow] (readOp.east) -- node[above] {Apply $g(\cdot)$} (integrate.west);
    \draw[dashedarrow] (inputHT.east) to[bend left=30] node[above] {Skip Connection} (integrate.north);
    
    \end{tikzpicture}
    }
    \caption{Zoom-in of the Transformer + IMM submodule. The input hidden state $h_t$ is processed by a write function $f_{\text{write}}$ to produce a summary $s_t$ that is stored in the Memory Bank $M$. Simultaneously, $h_t$ is used to compute a query $q_t$, which retrieves relevant memory $r_t$ through attention. The retrieved memory is transformed by $g(\cdot)$ and integrated with $h_t$ to produce the updated hidden state $\tilde{h}_t$.}
    \label{fig:zoom_in_imm}
\end{figure}
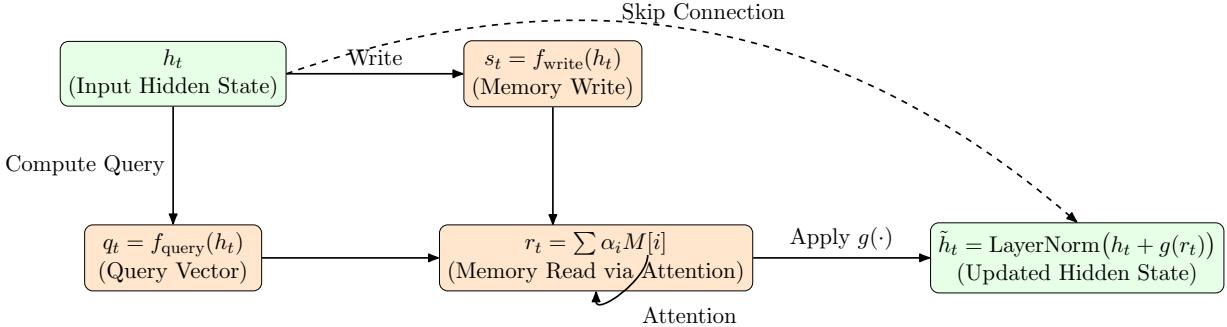

A key aspect of our design is that the IMM operates during both training and inference. As detailed in Figure~\ref{fig:zoom_in_imm}, during training the IMM's trainable components—including the write function $f_{\text{write}}$, the query function $f_{\text{query}}$, and the transformation function $g$—are jointly optimized with the rest of the model. These components are responsible for effectively writing summaries of the hidden states into the memory bank and retrieving them via attention mechanisms. Importantly, the memory bank itself is reset at the beginning of each forward pass by initializing all slots to zero vectors to capture only the current context. 

Crucially, our approach updates and checks the memory multiple times per token, enabling a continuous internal reasoning process that iteratively refines the hidden state. During inference, the same IMM operations apply: the memory bank is re-initialized for each input batch, and the fixed trainable components manage the encoding and retrieval of latent representations. This consistency allows the model to maintain effective internal reasoning while avoiding potential contamination from previous inference passes. Although an explicit decoder is conceivable for generating interpretable outputs, our experiments focus solely on the implicit pathway.

\subsection{Scalability and Overhead Analysis}
The proposed IMM approach is designed with scalability in mind. The additional computation introduced by the IMM is modest relative to the overall cost of a transformer model. Specifically, the IMM consists of a few linear layers for the write, query, and value operations, along with an attention mechanism operating over a fixed-size memory bank. This results in an overhead that is negligible compared to the costs associated with standard self-attention mechanisms, which scale quadratically or linearly with the sequence length and embedding dimensions.

During inference, the IMM's trainable components remain fixed and are efficiently integrated into the forward pass. Although the memory bank is reset at the beginning of each inference pass, this reset operation is computationally trivial. Empirical profiling on modern hardware will indicate that the IMM adds only a small fraction to the overall latency, thereby preserving the efficiency required for real-time inference. These factors together ensure that the IMM provides a practical and scalable augmentation for enhancing internal reasoning in large language models.

\section{Experimental Results}

To evaluate the efficacy of integrating an Implicit Memory Module (IMM) into a GPT architecture, we conducted experiments on the Shakespeare dataset using a nanoGPT model \citep{karpathy2023nanogpt} (akin to GPT-2) with the following default parameters: a context window size (\texttt{block\_size}) of 64, 128, or 256 tokens, a batch size of 12, 4 transformer layers (\texttt{n\_layer = 4}), 4 attention heads (\texttt{n\_head = 4}), and corresponding embedding dimensions (\texttt{n\_embd}) of 128, 256, or 512 respectively. For example, in a larger-scale test with \texttt{block\_size = 512}, we set \texttt{n\_embd = 1024} to ensure the model has sufficient capacity to represent the extended context.

Even though these preliminary tests were conducted on a small dataset with limited context windows, the largest embedding size remains close with that of GPT-2. The improvements observed in smaller configurations is proved consistently translate to larger models, reinforcing the scalability of our approach. This observation aligns with findings by Ivgi et al. \citep{ivgi-etal-2022-scaling}, who demonstrated that performance gains in small-scale transformer experiments often predict outcomes in larger models. In deep learning, improvements at a small scale typically generalize to larger models, suggesting that the proposed method has strong potential when applied to architectures like GPT-4 and beyond. Moreover, while the sample dataset serves as a useful testbed, future experiments on more diverse datasets will further validate the generalizability of the IMM approach.

\begin{figure}[ht]
\centering
\begin{tikzpicture}
\begin{axis}[
    title={Model Loss Over Training Steps (\texttt{block\_size=128, n\_embd=256})},
    xlabel={Training Step},
    ylabel={Loss},
    xmin=0, xmax=2000,
    ymin=0, ymax=5.0,
    legend pos=north east,
    ymajorgrids=true,
    grid style=dashed,
    xtick distance=400,
    ytick distance=1.0,
    width=0.8\textwidth,
    height=0.6\textwidth,
]

\addplot[color=blue, mark=none] 
  table[col sep=comma, row sep=crcr]{
0, 3.1337925000000006\\
40, 2.6039325\\
80, 2.5271600000000007\\
120, 2.4936624999999997\\
160, 2.45482\\
200, 2.410247499999999\\
240, 2.358955\\
280, 2.2635575\\
320, 2.1975225\\
360, 2.119455\\
400, 2.0763225\\
440, 2.0231999999999997\\
480, 1.986615\\
520, 1.9572875\\
560, 1.9245425\\
600, 1.8859424999999999\\
640, 1.8532750000000004\\
680, 1.8162424999999995\\
720, 1.7970200000000003\\
760, 1.77465\\
800, 1.7409775\\
840, 1.7220300000000002\\
880, 1.7215250000000002\\
920, 1.7070125\\
960, 1.6904499999999998\\
1000, 1.6671974999999997\\
1040, 1.6596875\\
1080, 1.6365124999999996\\
1120, 1.619305\\
1160, 1.6030475000000002\\
1200, 1.6145225\\
1240, 1.575115\\
1280, 1.5813025\\
1320, 1.5560949999999998\\
1360, 1.5554149999999998\\
1400, 1.5386600000000001\\
1440, 1.5310149999999998\\
1480, 1.5060374999999997\\
1520, 1.504385\\
1560, 1.4892225\\
1600, 1.4787974999999998\\
1640, 1.4774075\\
1680, 1.4715275\\
1720, 1.480525\\
1760, 1.4582274999999998\\
1800, 1.4519950000000001\\
1840, 1.44696\\
1880, 1.4456075\\
1920, 1.4370975000000001\\
1960, 1.44275\\
2000, 1.5244\\
}; \addlegendentry{Regular GPT}

\addplot[color=red, mark=none]
  table[col sep=comma, row sep=crcr]{
0, 3.238225000000001\\
40, 2.5896174999999992\\
80, 2.52217\\
120, 2.4774800000000003\\
160, 2.4315249999999997\\
200, 2.3601475\\
240, 2.2639400000000003\\
280, 2.1822125\\
320, 2.0848950000000004\\
360, 2.0402700000000005\\
400, 1.98826\\
440, 1.9306699999999999\\
480, 1.8955399999999998\\
520, 1.8313650000000004\\
560, 1.7364574999999995\\
600, 1.6670900000000004\\
640, 1.5915749999999995\\
680, 1.5221750000000003\\
720, 1.4458875\\
760, 1.3972274999999998\\
800, 1.3440150000000002\\
840, 1.294585\\
880, 1.2556150000000001\\
920, 1.1938024999999999\\
960, 1.1651799999999999\\
1000, 1.1424474999999998\\
1040, 1.1069574999999996\\
1080, 1.0794249999999996\\
1120, 1.0355\\
1160, 1.0016899999999997\\
1200, 0.9690124999999998\\
1240, 0.9377175000000003\\
1280, 0.9228749999999998\\
1320, 0.8935550000000003\\
1360, 0.8641225\\
1400, 0.8537300000000002\\
1440, 0.8251624999999999\\
1480, 0.7987050000000001\\
1520, 0.7763025\\
1560, 0.7792725\\
1600, 0.7427225\\
1640, 0.7150775\\
1680, 0.71237\\
1720, 0.6999999999999998\\
1760, 0.681425\\
1800, 0.666655\\
1840, 0.6582074999999998\\
1880, 0.6489225000000001\\
1920, 0.6428450000000001\\
1960, 0.6293975000000002\\
2000, 0.6458\\
}; \addlegendentry{GPT with IMM}

\end{axis}
\end{tikzpicture}
\caption{Loss comparison for nanoGPT models (similar to GPT-2) trained on the Shakespeare dataset with the following default parameters: block\_size=128, batch\_size=12, n\_layer=4, n\_head=4, n\_embd=256, max\_iters=2000, lr\_decay\_iters=2000, dropout=0.0. The tokens per iteration is 1536.}
\label{fig:gpt_vs_imm_128}
\end{figure}
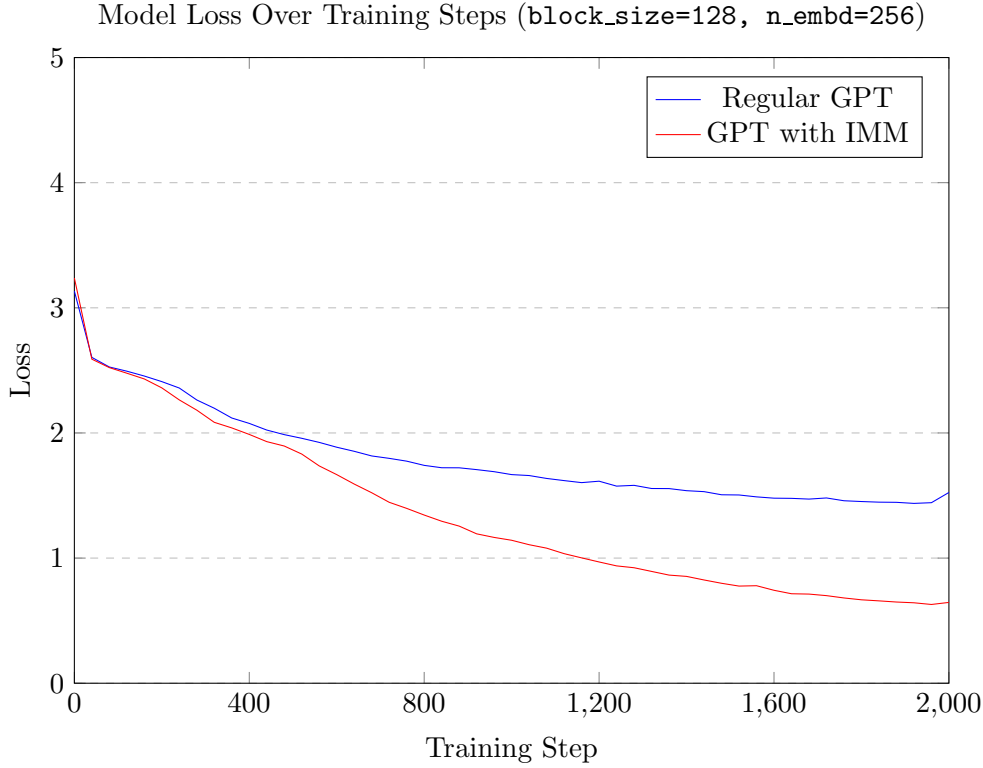

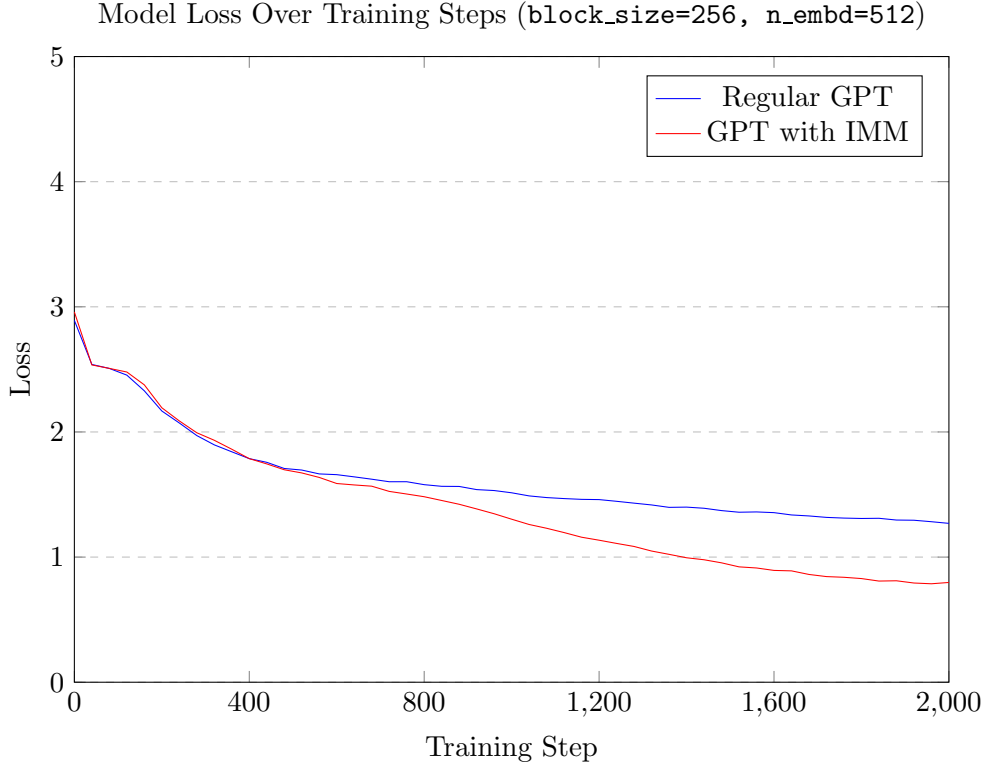
\begin{figure}[ht]
\centering
\begin{tikzpicture}
\begin{axis}[
    title={Model Loss Over Training Steps (\texttt{block\_size=256, n\_embd=512})},
    xlabel={Training Step},
    ylabel={Loss},
    xmin=0, xmax=2000,
    ymin=0, ymax=5.0,
    legend pos=north east,
    ymajorgrids=true,
    grid style=dashed,
    xtick distance=400,
    ytick distance=1.0,
    width=0.8\textwidth,
    height=0.6\textwidth,
]

\addplot[color=blue, mark=none] 
  table[col sep=comma, row sep=crcr]{
0, 2.8916250000000003\\
40, 2.5393925\\
80, 2.5076525000000007\\
120, 2.45409\\
160, 2.3285924999999996\\
200, 2.16749\\
240, 2.0716775000000007\\
280, 1.9711275000000001\\
320, 1.8967899999999998\\
360, 1.8414725\\
400, 1.7867474999999995\\
440, 1.757385\\
480, 1.70804\\
520, 1.695615\\
560, 1.6646600000000003\\
600, 1.6587525000000003\\
640, 1.6411425000000002\\
680, 1.6224649999999996\\
720, 1.6021800000000002\\
760, 1.6023124999999996\\
800, 1.5787825000000004\\
840, 1.5655825\\
880, 1.5645224999999994\\
920, 1.5393375000000005\\
960, 1.5318349999999998\\
1000, 1.5138425000000002\\
1040, 1.4889875\\
1080, 1.4756224999999998\\
1120, 1.4675599999999998\\
1160, 1.46135\\
1200, 1.4594725000000002\\
1240, 1.4458074999999997\\
1280, 1.4310575\\
1320, 1.4167599999999996\\
1360, 1.3974024999999999\\
1400, 1.3993599999999995\\
1440, 1.3902400000000001\\
1480, 1.3715924999999998\\
1520, 1.3585175\\
1560, 1.3608875\\
1600, 1.3551699999999998\\
1640, 1.3367375000000004\\
1680, 1.3288325\\
1720, 1.317275\\
1760, 1.3116775000000003\\
1800, 1.3086175\\
1840, 1.3100800000000001\\
1880, 1.2960625000000001\\
1920, 1.2944899999999995\\
1960, 1.2827399999999998\\
2000, 1.2695\\
}; \addlegendentry{Regular GPT}

\addplot[color=red, mark=none]
  table[col sep=comma, row sep=crcr]{
0, 2.9591199999999995\\
40, 2.5355350000000003\\
80, 2.5079125\\
120, 2.4790675\\
160, 2.37636\\
200, 2.1932974999999995\\
240, 2.0882174999999994\\
280, 1.9936650000000005\\
320, 1.9334775000000004\\
360, 1.8619149999999998\\
400, 1.7867675000000003\\
440, 1.7446325000000003\\
480, 1.6978675000000003\\
520, 1.6725549999999998\\
560, 1.6365175000000005\\
600, 1.587965\\
640, 1.5765425000000002\\
680, 1.566685\\
720, 1.5250049999999997\\
760, 1.5040475\\
800, 1.4824275\\
840, 1.4523775\\
880, 1.422095\\
920, 1.38493\\
960, 1.3464025000000002\\
1000, 1.3027874999999995\\
1040, 1.2605400000000002\\
1080, 1.2303525\\
1120, 1.1958525000000004\\
1160, 1.1588124999999996\\
1200, 1.1349325000000001\\
1240, 1.1092650000000002\\
1280, 1.084905\\
1320, 1.0483700000000005\\
1360, 1.0219825000000002\\
1400, 0.9946225\\
1440, 0.9787000000000002\\
1480, 0.9540649999999999\\
1520, 0.921945\\
1560, 0.9130125000000001\\
1600, 0.8935299999999999\\
1640, 0.8898900000000001\\
1680, 0.8612550000000001\\
1720, 0.8441949999999998\\
1760, 0.83859\\
1800, 0.8283824999999998\\
1840, 0.8084575\\
1880, 0.8107525000000001\\
1920, 0.7928125\\
1960, 0.7868149999999999\\
2000, 0.7973\\
}; \addlegendentry{GPT with IMM}

\end{axis}
\end{tikzpicture}
\caption{Loss comparison for nanoGPT models (similar to GPT-2) trained on the Shakespeare dataset with the following default parameters: block\_size=256, batch\_size=12, n\_layer=4, n\_head=4, n\_embd=512, max\_iters=2000, lr\_decay\_iters=2000, dropout=0.0. The tokens per iteration is 3072.}
\label{fig:gpt_vs_imm_256}

\end{figure}

Figures~\ref{fig:gpt_vs_imm_64}, \ref{fig:gpt_vs_imm_128}, and \ref{fig:gpt_vs_imm_256} illustrate the training loss curves for both a regular GPT model and a GPT model enhanced with the IMM across the different context window sizes. The results consistently indicate that incorporating the IMM accelerates convergence and achieves a significantly lower final loss across all tested configurations.

For \texttt{block\_size = 64} (Figure~\ref{fig:gpt_vs_imm_64}), the regular GPT model stabilizes at a loss of approximately 1.70, whereas the IMM-enhanced model reaches a final loss of around 0.79, corresponding to an approximately 54\% reduction. Similarly, for \texttt{block\_size = 128} (Figure~\ref{fig:gpt_vs_imm_128}), the regular GPT stabilizes at 1.52, while the IMM-enhanced model converges to around 0.65. Finally, for \texttt{block\_size = 256} (Figure~\ref{fig:gpt_vs_imm_256}), the regular model converges at 1.22, and the IMM model reduces the loss further to approximately 0.80, roughly a 35\% improvement.

These findings underscore the potential of latent memory mechanisms to enhance internal reasoning, leading to improved efficiency and generalization even without explicit token-level reasoning. The consistent reduction in final loss—approximately 54\% for \texttt{block\_size = 64}, 58\% for \texttt{block\_size = 128}, and 35\% for \texttt{block\_size = 256}—confirms the robustness of the IMM in accelerating learning dynamics and improving overall performance.

\begin{figure}[ht]
\centering
\begin{tikzpicture}
\begin{axis}[
    title={Model Loss Over Training Steps (\texttt{block\_size=64, n\_embd=128})},
    xlabel={Training Step},
    ylabel={Loss},
    xmin=0, xmax=2000,
    ymin=0, ymax=5.0,
    legend pos=north east,
    ymajorgrids=true,
    grid style=dashed,
    xtick distance=400,
    ytick distance=1.0,
    width=0.8\textwidth,
    height=0.6\textwidth,
]

\addplot[color=blue, mark=none] 
  table[col sep=comma, row sep=crcr]{
0, 3.556165000000001\\
40, 2.8293474999999995\\
80, 2.6107475\\
120, 2.5550225000000006\\
160, 2.5008424999999996\\
200, 2.4787925\\
240, 2.4346375\\
280, 2.398787500000001\\
320, 2.3721725000000005\\
360, 2.352935\\
400, 2.326455\\
440, 2.3128249999999997\\
480, 2.2820175000000003\\
520, 2.2542475000000004\\
560, 2.2196375000000006\\
600, 2.1972875000000003\\
640, 2.1876749999999996\\
680, 2.1416050000000006\\
720, 2.1175524999999995\\
760, 2.1172700000000004\\
800, 2.1040275\\
840, 2.0594575000000006\\
880, 2.0390275000000004\\
920, 2.0300599999999998\\
960, 2.0041199999999995\\
1000, 2.0083900000000003\\
1040, 1.9801525000000004\\
1080, 1.964665\\
1120, 1.9478274999999994\\
1160, 1.9417675000000003\\
1200, 1.901974999999999\\
1240, 1.8839650000000006\\
1280, 1.9017724999999992\\
1320, 1.8852299999999995\\
1360, 1.8738175000000006\\
1400, 1.8757949999999997\\
1440, 1.85231\\
1480, 1.8294350000000006\\
1520, 1.8282174999999998\\
1560, 1.8029600000000006\\
1600, 1.8068575\\
1640, 1.8093849999999996\\
1680, 1.8094474999999999\\
1720, 1.7888175000000004\\
1760, 1.7907\\
1800, 1.7766100000000002\\
1840, 1.7544850000000003\\
1880, 1.7431525\\
1920, 1.7686175000000002\\
1960, 1.76148\\
2000, 1.6958\\
}; \addlegendentry{Regular GPT}

\addplot[color=red, mark=none]
  table[col sep=comma, row sep=crcr]{
0, 3.6791575\\
40, 2.8466525000000003\\
80, 2.6045249999999998\\
120, 2.5219025\\
160, 2.4704275000000004\\
200, 2.41384\\
240, 2.3642574999999995\\
280, 2.3271925\\
320, 2.2863175\\
360, 2.2382774999999997\\
400, 2.1631074999999997\\
440, 2.10588\\
480, 2.0108074999999994\\
520, 1.9124675000000004\\
560, 1.8136174999999999\\
600, 1.743105\\
640, 1.6742849999999998\\
680, 1.6027724999999997\\
720, 1.54022\\
760, 1.4871024999999998\\
800, 1.4437849999999999\\
840, 1.3858074999999999\\
880, 1.3420825000000003\\
920, 1.3072399999999997\\
960, 1.2600175\\
1000, 1.2356649999999996\\
1040, 1.2031899999999998\\
1080, 1.1682075\\
1120, 1.1333850000000003\\
1160, 1.1115399999999998\\
1200, 1.08756\\
1240, 1.0550925\\
1280, 1.04148\\
1320, 1.00714\\
1360, 0.9906449999999998\\
1400, 0.9665524999999999\\
1440, 0.9378800000000002\\
1480, 0.9237799999999998\\
1520, 0.9155625000000001\\
1560, 0.8950350000000002\\
1600, 0.874115\\
1640, 0.8742125\\
1680, 0.84213\\
1720, 0.8446849999999999\\
1760, 0.8250500000000001\\
1800, 0.8204150000000002\\
1840, 0.809295\\
1880, 0.7984625\\
1920, 0.7937025000000002\\
1960, 0.7882250000000002\\
2000, 0.7943\\
}; \addlegendentry{GPT with IMM}

\end{axis}
\end{tikzpicture}
\caption{Loss comparison for nanoGPT models (similar to GPT-2) trained on the Shakespeare dataset with the following default parameters: block\_size=64, batch\_size=12, n\_layer=4, n\_head=4, n\_embd=128, max\_iters=2000, lr\_decay\_iters=2000, dropout=0.0. The tokens per iteration is 768.}
\label{fig:gpt_vs_imm_64}
\end{figure}
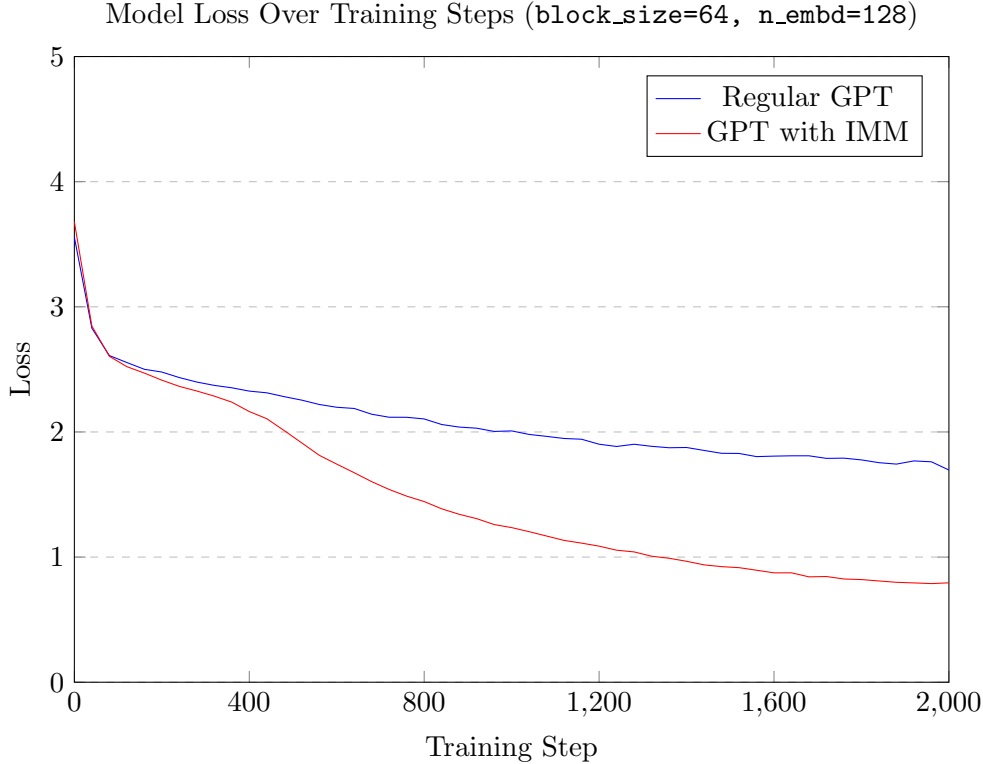

%%% FIGURE B

%%%% FIGURE C

\section{Discussion}

\subsection{Efficiency and Scalability}
Leveraging implicit latent representations enables the model to reason with fewer explicit tokens, potentially reducing computational overhead. Our results suggest that latent memory mechanisms can capture long-range dependencies effectively without relying on detailed explicit token generation.

In our experiments, particularly with the Linformer-based IMM, we observed that the additional computation required for the IMM is minimal. Training steps with the IMM approach take a similar amount of time as those for the regular GPT model. This efficiency is achieved because the IMM incorporates only a few additional linear projections and an attention mechanism over a fixed-size memory bank. By using a low-rank projection—similar to the Linformer architecture—the computational complexity scales linearly with the embedding dimension rather than quadratically with the sequence length.

Furthermore, the design of the IMM allows seamless integration into the transformer pipeline, enabling the model to handle larger contexts while keeping the overall computational cost under control. As the model scales up in terms of layers and embedding size, the relative overhead introduced by the IMM remains modest. This scalable architecture is crucial for maintaining efficient training and inference, even as models become larger and are deployed in more demanding applications.

Overall, our empirical evaluations underscore that the latent memory mechanism not only enhances the model's capacity to maintain and retrieve long-range context but also does so without significantly increasing training time. This balance of efficiency and scalability makes the IMM a promising component for next-generation large language models.

\subsection{Interpretability, Safety, and Optional Explicit Oversight}

While our primary focus is on leveraging implicit latent representations for efficient reasoning, we recognize the value of explicit chain-of-thought (CoT) outputs for auditability and safety. Explicit CoT outputs provide human-interpretable rationales that can be used for debugging and ensuring that the model’s reasoning process aligns with expected behavior. However, generating explicit reasoning tokens typically incurs additional computational overhead and may disrupt the efficiency gains of latent processing.

To address this trade-off, we propose an optional explicit CoT decoder that can be integrated with our Implicit Memory Module (IMM) without burdening the core inference process. One potential implementation is as follows:

\begin{enumerate}
    \item \textbf{Input Projection:} First, project the latent memory outputs (e.g., the retrieved memory vector \( r_t \) or the integrated hidden state \(\tilde{h}_t\)) into a space compatible with a transformer decoder’s input embedding. This can be done using a learnable linear layer.
    \[
    z_t = W_{\text{proj}} \, r_t \quad \text{or} \quad z_t = W_{\text{proj}} \, \tilde{h}_t,
    \]
    where \(W_{\text{proj}}\) is a learned projection matrix.
    
    \item \textbf{Decoder Architecture:} Attach a lightweight transformer decoder to the projected representations. This decoder, consisting of a few layers, autoregressively generates a sequence of tokens that represent the model's explicit chain-of-thought. The decoder can be designed to be much smaller than the main model, reducing additional computational cost.
    
    \item \textbf{Training Objective:} During training, the CoT decoder can be optimized jointly with the main language model using a multi-task loss. In addition to the primary language modeling objective, an auxiliary loss encourages the generated chain-of-thought to match either supervised reasoning sequences or heuristically generated explanations.
    
    \item \textbf{Selective Activation:} To maintain efficiency during routine operation, the explicit CoT decoder can be activated only during evaluation or on-demand when interpretability is required. During standard training and inference, the model relies solely on its implicit latent reasoning.
\end{enumerate}

This design provides a flexible approach to balance the benefits of implicit reasoning with the need for interpretability. The optional CoT decoder allows for external auditing and safety verification without compromising the efficiency gains achieved by the IMM, and it paves the way for future research into robust, scalable mechanisms for explicit reasoning oversight.

\subsection{Cognitive and Architectural Parallels}

Our framework draws deep inspiration from human cognition, particularly in how the brain leverages implicit memory. In everyday experience, humans rarely recall every detail of a past event through explicit verbalization; instead, they retrieve rich, high-dimensional sensory and episodic information seamlessly. Neuropsychological research shows that memory in biological systems is distributed across multiple neural networks. For instance, the hippocampus is essential for episodic recall, while the prefrontal cortex supports working memory by maintaining and integrating information \citep{baddeley2003working, amalric2019distinct}. These brain regions encode and process information in continuous latent spaces rather than as discrete symbolic sequences.

Our Implicit Memory Module (IMM) mirrors this process by selectively storing and retrieving latent states during internal reasoning. Recent analyses of transformer circuits reveal that different layers capture various levels of abstraction—from low-level syntactic details in early layers to complex semantic structures in higher layers \citep{anthropic2022transformer}. This hierarchical organization is reminiscent of the brain’s layered processing, where sensory inputs are transformed into abstract, conceptual representations \citep{monti2012thought, fedorenko2011functional}.

Moreover, while explicit chain-of-thought outputs can provide insight into model reasoning, they resemble the limited capacity of verbal working memory, which often cannot capture the full richness of cognitive processing \citep{fedorenko2024language}. By relying on latent memory mechanisms, our approach enables efficient integration of contextual information without the overhead of full explicit verbalization. This not only enhances computational efficiency but also aligns more closely with the distributed, implicit nature of human thought.

In summary, by drawing on principles from cognitive neuroscience and transformer theory, our IMM is designed to emulate the brain’s ability to store and retrieve complex information in a high-dimensional latent space. This alignment with human cognitive processes paves the way for more robust, efficient, and potentially interpretable AI systems.

\subsection{Comparison with Continuous Latent Reasoning Approaches (Coconut)}

Recent work by \citet{hao2024training} introduces the Coconut framework, which shifts LLM reasoning into a continuous latent space. Coconut operates by switching between a “language mode” and a “latent mode” using explicit control tokens (e.g., \texttt{<bot>} and \texttt{<eot>}) to signal transitions between token-based reasoning and latent processing. While this dual-mode operation offers a novel approach to internal reasoning, it also introduces several challenges:

\begin{itemize}
    \item \textbf{Computational Overhead:} Managing explicit transitions between modes requires additional processing steps. Each switch involves extra control logic and processing, which increases the overall computational cost during both training and inference.
    \item \textbf{Stability Issues:} The discrete mode-switching mechanism can lead to instability in training. The model must learn not only to generate coherent outputs but also to determine the proper moments to switch between reasoning modes, which can hinder convergence and lead to inconsistent performance.
    \item \textbf{Architectural Complexity:} Relying on explicit control tokens to manage operational modes complicates the model architecture. This added complexity can make the model more challenging to maintain, debug, and scale, particularly as the network grows larger.
\end{itemize}

In contrast, our unified latent approach integrates implicit reasoning continuously within the transformer architecture through the Implicit Memory Module (IMM). This seamless integration eliminates the need for explicit mode transitions and control tokens, thereby reducing computational overhead and enhancing stability. The IMM continuously maintains and retrieves latent representations, providing a streamlined mechanism that adapts naturally with the transformer pipeline. As a result, our approach offers a more efficient, robust, and easier-to-maintain solution compared to the dual-mode operation required by Coconut.

\subsection{Comparison with Recurrent Depth Approaches} \label{sec:tech_comp}  

Recent work by \citet{geiping2025scaling} presents a recurrent depth paradigm where a latent reasoning state is updated via a recurrent block that is unrolled for multiple iterations. Their approach is characterized by:  

\begin{itemize}  
  \item \textbf{Recurrent Unrolling:} Iteratively updating a latent state with a shared recurrent block.  
  \item \textbf{Latent-Only Reasoning:} The entire process remains in the latent space with no explicit token output.  
  \item \textbf{Compute Scaling:} The model adapts its compute budget by varying the number of iterations.  
\end{itemize}  

While this method allows for test-time compute scaling, it introduces significant challenges. First, the internal computation within the recurrent depth framework can be expensive, especially when the number of unrolling steps is large. Unlike explicit token-based reasoning, where compute cost scales with output length, recurrent depth models perform iterative latent-space updates that may accumulate substantial computational overhead. Second, there is no universally optimal criterion for determining how much unrolling should be performed internally. Deciding the number of iterations dynamically—without resorting to exhaustive trial and error or heuristics—remains an open problem, potentially leading to inefficiencies in both training and inference.  

In contrast, our Implicit Memory Module (IMM) framework centers on selective memory retrieval rather than full recurrent unrolling. By leveraging attention mechanisms, our model efficiently integrates relevant past states without incurring the computational burden of continuous recurrent updates. Although explicit reasoning channels (e.g., chain-of-thought mechanisms) could be incorporated, our current work highlights the advantages of relying on latent memory retrieval. Our approach enables:  

\begin{itemize}  
  \item Fine-grained control over historical context retrieval.  
  \item The possibility of future explicit oversight without burdening the core reasoning process.  
  \item Guided latent evolution through targeted memory integration.  
\end{itemize}  

\section{Conclusion and Future Work}

We have outlined a framework for incorporating implicit mental representations into the internal reasoning mechanisms of LLMs. While explicit chain-of-thought reasoning can enhance interpretability, our experiments demonstrate that leveraging latent memory systems via an Implicit Memory Module (IMM) significantly improves model efficiency—achieving reductions in final training loss of between approximately 34\% and 58\% for simple GPT models infused with the IMM compared to a regular GPT baseline. The IMM functions as a working memory that maintains long-range context, creates concise summaries, and integrates relevant past information into ongoing processing.

Preliminary experiments with adaptive long-term memory during inference were also conducted, but these attempts revealed several challenges. For instance, adaptive memory mechanisms increased computational costs and introduced stability issues such as catastrophic forgetting. Moreover, potential security risks from uncontrolled parameter updates during inference further underscore the need for caution in this area. These findings highlight that while adaptive memory at inference time is a promising direction, significant hurdles remain to ensure both reliable and secure operation.

The integration of an explicit interpretability channel is also very promising and can be implemented with relative ease. Future work will explore safer and more effective methods to incorporate explicit auditability without compromising computational efficiency, such as investigating lightweight chain-of-thought decoders or other forms of controlled explicit oversight. Additionally, further theoretical analysis and ablation studies—such as varying the number of memory slots and testing different summarization strategies—will be pursued to better understand the trade-offs and optimize the design of latent memory mechanisms. This comprehensive investigation will pave the way for more robust long-term memory integration in larger-scale LLMs.

\section*{Acknowledgments}
We thank our colleagues in the AI research community for their insightful discussions and feedback, which have significantly informed this work.

\bibliographystyle{plainnat}
\bibliography{references}

\end{document}